A Deep Learning Based Chatbot for Campus Psychological Therapy

# CATALOG

# *Evebot*: A Deep Learning Based Chatbot System for Campus Psychological Therapy

Junjie Yin*, Zixun Chen, Kelai Zhou, Chongyuan Yu

Guangdong Country Garden School, Foshan, China

{oscarfantasywork@outlook.com }

**Abstract:**

In this paper, we propose *Evebot*, an innovative, sequence to sequence (Seq2seq) based, fully generative conversational system for the diagnosis of negative emotions and prevention of depression through positively suggestive responses. The system consists of an assemble of deep-learning based model, including Bi-LSTM based model for detecting negative emotion of users and obtaining psychological counselling related corpus for training the chatbot, anti-language sequence to sequence neural network, and maximum mutual information (MMI) model. As adolescents are reluctant to show their negative emotion in physical interaction, traditional method of emotion analysis and comforting methods may not work. Therefore, this system put emphasis on using virtual platform to detect signs of depression or anxiety, channel adolescents' stress and mood, and thus prevent the emergence of mental illness. We launched the integrated chatbot system onto an online platform for real-world campus application. Through a one-month user study, we observe better results in the increase in positivity than other public chatbot in the control group.

**Keywords**: Chatbot, Conversational System, NLP, Psychological Therapy, Deep Learning

## 1 Introduction

With increasing pressure from schools nowadays, adolescents are prone to psychological illnesses, including mild mental disorders, depressions, anxiety, and possibly suicidal behavior, because of their immaturity in both emotion and spirit (Hawton, 2012). Therefore, it is a priority of psychologists to handle adolescent stress (Compas, 1993), as a means of preventing mental illness. However, most students with stress or mental illnesses are often reluctant or unwilling to share their true feelings to other people, and it is even more unlikely that they will voluntarily seek for psychological assistance. Nevertheless, the advent of the Internet has brought great potential to addressing these problems. Many students are turning to the Internet to release their negative feelings, and it is shown to have positive effects on loneliness, depression, and stress (Shaw, 2004). It is then reasonable that a chatbot, or a chatting system based on artificial intelligence, can serve as a "virtual friend" to release the negative emotion of students, as it allows students to express their true feeling that cannot be otherwise expressed in real life. In light of Craig G. Rogers's concept of "Client-centered therapy", non-medical staff can also perform psychological counseling on mild mental disorders or illnesses with proper training (Rogers, 1995). This had laid a foundation for the application of a chatbot system in psychotherapy, as it can offer support for adolescents with mild mental disorders, helping the release of their negative emotions.

Recently, Liu et al. proposed a chatbot named PAL, which can answer non-obstructive psychological domain-specific questions. They achieve this through collecting question-and-answer (Q&A) pairs into a knowledge base and retrieve the answer by matching with semantics (Liu, 2013). This approach can perform in an ideal setting, but in most real-life scenarios, users would normally chat with the bot to tell their misfortune instead of asking specific psychological questions. In addition, giving solutions to one psychological problem may be not enough to help depressive users. They also need the feeling of being cared as their feelings are being listened, understood, and comforted. Overall, chatbots that answer psychological domain-specific questions is inadequate in releasing user's negative emotions and feelings as they are unable comprehend and sooth their emotion.

In this paper, we propose a student-specific, deep learning chatbot system 'Evebot', capable of diagnosing students negative, depressive, and anxious emotion during chatting, and acting as a psychological therapist and virtual friend throughout the conversation to channel their negative emotion by comforting their feelings, instilling positive emotion, and offering solutions. To our knowledge, Huang et al. has created a chatbot similar in function to our proposed approach, which aims specifically to release adolescents stress. However, the response generation models proposed by Huang et al. is through selecting answers from local knowledge base and from Chinese research engine website Baidu. The response is fixed and may not answer the user's query appropriately. Comparatively, our chatbot system is based on state-of-the-art deep learning models, including bidirectional recurrent neural networks and sequence-to-sequence models, to produce natural language responses and generate responses specific to each student's query and learning the student's tone as each conversation continues.

The paper consists of the following: We review and analyze the related work in Section 2 and demonstrate the overall structure of our Evebot in Section 3. Then we present three crucial parts, including negative emotion detection, psychological corpora retrieval, and response generation in Sections 4 to 6 respectively. Next, Section 7 evaluates the effectiveness of our chatbot through experiment on students. Finally, conclusions and expectations are present in Section 8.

## 2 Related Work

Chatbot (Chatterbot): Chatbots are based on Q&A systems, which retrieve the matching question-answering pair from knowledge bases to respond to user's response (Shilin Ding et al., 2008). Q&A systems can also respond by searching through Web Pages and related documents (Lin & Katz, 2003). As a special kind of Q&A systems, chatbot focus more on being anthropomorphic. Its purpose is to talk to users as if it is a real human. While current chatbot system have not emulate people talking to an extent that is even close to an adolescent, several approaches have managed to make significant progress. Alice and Eliza are two of the most well-known examples, which engage with users to either act like a psychotherapist to understand user's response or a friendly friend that start conversation (Weizenbaum, 1966; Liao, 2005). Inspired by these early chatbots, current chatbots hold a wide range of application, including in e-commerce, medical heath care, and intelligent tutoring systems.

Natural Language Processing: Recurrent neural network and convolutional neural networks are predominant in handling NLP tasks. A novel architecture (Alexis Conneau et al., 2016) by using deep layers that would typically see in computer vision to perform text processing, and have improved current classification tasks. They have concluded that with more depth, the performance of the model will be enhanced. This is the first time deep convolutional networks are applied into NLP and offered an insight into how this can improve related tasks. Another area of application is in opinion mining, also known as sentiment analysis, which is an essential approach to analyze data. NLP techniques are reviewed for text preprocessing, opinion mining approaches are investigated for different situations. (Sun, Shiliang, Chen Luo, and Junyu Chen., 2017). NLP techniques can also be used in order to learn, understand and produce human language. Real world applications include making spoken dialogue systems and carrying out social media mining. (Hirschberg & Manning, 2015.)

Speech and Dialogue: The Dynamic Memory Network (DMN) is a neural network architecture that processes input sequences, creates episodic memories (memories relevant to the self) and produces relevant answers. This can be trained to thoroughly on many types of tasks, including sequence modeling for part-of-speech tagging (WSJ-PTB). (Kumar, Ankit, et al., 2016) Common Natural Language Generation (NLG) uses employ rules and heuristics, causing the responses generally to be monotonous and rigid. Wen and his peers have developed Long Short-Term Memory (LSTM) structure-based generator. Using cross entropy loss function, the LSTM based generator could learn from unaligned data and optimized the sentence planning. This approach proved to have fewer heuristics and improved performance than the previous methods. (Wen et al., 2015). Supervised learning NLP techniques can now automatically detect hate speech, though with limitations. There is also a possibility of constructing a benchmark dataset for hate speech recognition. (Schmidt, Anna, and Michael Wiegand., 2017).

Textural Sentiment Diagnosis: The most common approach of emotion diagnosis is through searching keywords (Subasic, 2001; Balahur, 2011). Such method can achieve high accuracy, but only when the response explicitly contains emotional words or phrases, which is rare in real-life conversation. Deep learning-based approaches have made progress consider its ability in image processing and speech. Convolutional neural network (Wang, 2016) and LSTM network (Gupta, 2017) are used and prove to be effective in classifying emotion.

Application in Psychology: Motivational Interviewing (MI) is a style of psychological counseling that focuses on changing the client's behavior and is often used when the client is addicted to substances. Two NLP systems based on DSF and RNN are used in this study to automatically code MI sessions, and predictions from these models are compared to human ratings from a large sample. The DSF model performs slightly better at utterance- and session-level agreement than the NLP model, as the RNN model is trained with more formal sources, but poor agreement is observed with both models in some cases. NLP models have the potential to allow clinical supervision to be a practice (Tanana, Michael, et al., 2016).

## 3 System Overview

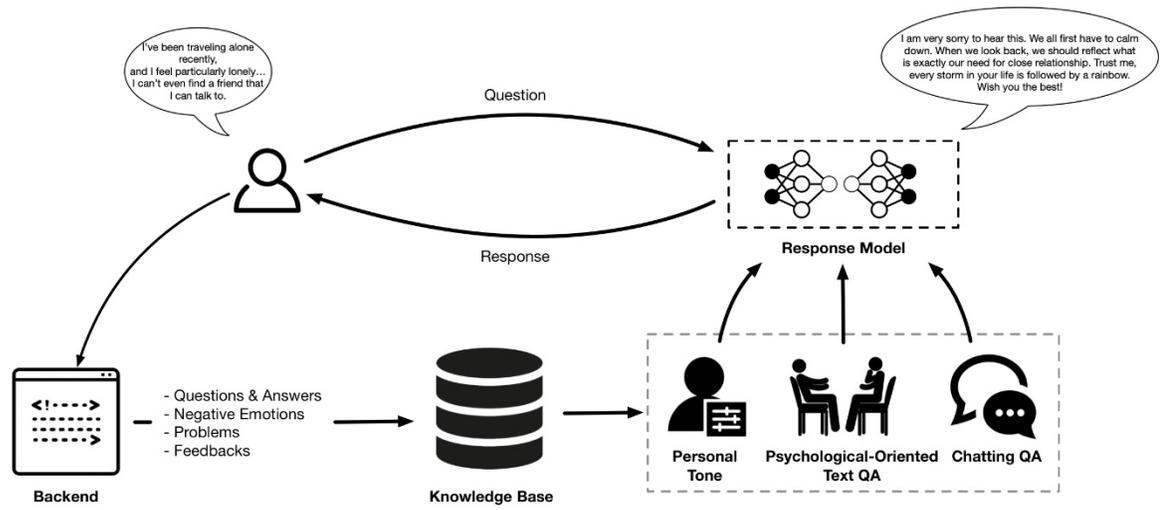

*Figure 3.1: System Overview*

Our system framework is a combination of independent sub-models that can work together effectively. Figure 3.1 shows its framework. The user chats with Evebot on an online social media platform "WeChat[1]". These systems consist of two chatbots, one sentiment analysis model, and one classifier model. The response model includes the casual chatting chatbot and psychological counselling chatbot. Determined by user's responses, the response model decide to use which chatbot to answer. Both models take a sequence of words as input and another sequence of words as output. The response models are decided through the classifier model, which take a sentence as input and give a label of either casual or mental-related to that sentence.

The sentiment analysis model is responsible for detecting the emotion of the user over the entire period of conversation. Each user's response will be scored from 0 to 1 to give an evaluation on user's emotion, 0 denoted as the most negative and 1 denoted as the most positive.

When the user gives its response back to the response model, the backend of the system will also record the information and store it into knowledge base, which is fit to the model in the form of Q&A pair.

---

[1] WeChat: a free mobile application by Tencent providing online communication service. Tencent claims 94% of all smartphones in China has WeChat installed, with Monthly Active Users reaching 806 million in Q2 2016.

## 4 Detecting Mood-Sentiment Analysis

### 4.1 Problem Definition

Following our identification of adolescent stress and the following psychological distresses, it is crucial for our chatbot model to detect the stress effectively through emotional expressions in natural language. In order to efficiently accomplish this, we quantitatively score the expression of human sentiments on a scale of negative (0) to positive (1). Then we use data collected from our survey combined with a Bidirectional LSTM-based RNN to produce judgements.

### 4.2 Collecting Data

As psychological counseling is a medical field where patients' personal data are sensitive and usually hidden from public access, we decided to first-handedly obtain data from a survey handed out to students in our high school in order to train a sentiment classifier.

We first made a visit to the school psychological consultation center and obtained several lists of psychological diagnosis surveys that psychologists commonly instruct their patients to complete, including the SCL-90, SAS/SDS, MHT, MBTI and PHQ9.

We then selected 10 questions from all these surveys that can cover the most varied responses, and paraphrase each of these questions to be more abstract in nature. This not only allows students answering our survey to respond without clear limitations, but also eliminates potential response bias in that students have less chance of perceiving the survey as concerning their mental health, and therefore will answer truthfully. We also added five confounding questions for the survey, selected from the "Sorting Hat" survey on Pottermore.com, to further decrease the chance of students perceiving our survey as mental health related. Responses to these questions will not be used as training data. Table 3.1 gives two examples of a survey question and one example of a confounding question.

| |
|---|
| **Survey Question: During in group discussion, if someone holds different opinion, what would you do? (Paraphrase from PHQ-9)** |
| 1 我会通过沟通使他们会肯定我的意见 |
| (I would persuade them through communication) |
| 1 别人的想法应该有创意，然后我们可以进行修改变得更加可行 |
| (Others need to have creative idea, and we should rectify the idea to be more plausible) |
| 0 他有什么问题啊… |
| (What's the problem with him…) |
| **Survey Question: Soon is your birthday, if you could do anything, what would that be? (Paraphrase from SCL-90)** |
| 0 没啥好干的吧 |
| (Well, there is nothing to do) |
| 1 享受与家人一起旅行，生活 |
| ([I would] spend time with my family travelling and enjoying life) |
| 1 去看 Ed Sheeran 的演唱会 |
| ([I would] go to Ed Sheeran's concert) |
| **Confounding Question: I would like to discover…** |
| N/A 一个最强大脑 |
| (A powerful mind) |
| N/A 能治愈任何疾病的植物 |
| (A plant that can cure any disease) |
| N/A 一个只对我忠诚的助手 |
| (An assistant loyal only to me) |

*Table 4.1:* Examples of the Results of Sentiment Analysis

We set out to survey a random sample of high schoolers in our high school, all participating in the AP program and the AP examination. As this program is commonly considered the most stressful in our school, and the limited time from the AP examination is likely to drive students to study harder, students should experience a considerable amount of stress. Under this stress, students are more likely to provide meaningful (more psychology-related) responses to the survey questions, balancing out the non-related responses, and this effect is indeed observed in the collected data. As we are not conducting a study to capture any population parameter accurately by this survey, there is no concern of potential nonresponse bias.

There is a total of 99 responses to the survey, and with 10 questions for each survey, we have a total data size of 990 individual responses to questions.

We then separate the survey responses into two categories, "Negative Emotion" and "Positive Emotion". As we are only interested in negative emotions, neutral responses are put into the Positive category. It is not unreliable to separate responses by any machine classification algorithm, as accuracy may not be high enough, so we classify all 990 responses by hand.

After this classification, the data collected can be represented by:

$$R \in \{r_1, r_2, …, r_i\} \tag{1}$$

Where $R$ denotes the raw data being collected, and $r_i$ is response at i[th] position with a label as either positive or negative.

Now we can utilize the data in training a machine learning based classifier.

### 4.3 Bi-LSTM-based Mood Detection

Recurrent Neural Networks (RNNs) are a strong tool for sequence data including speech, handwriting recognition and machine translation. Because of the inherent vanishing gradient problem, LSTM network has become a major architecture of RNN. Several studies have compared the performance of different RNN structures in constructing an NLP-based binary classification model and found that while GRU models has a high accuracy, LSTM models are more confident in their decision.

However, LSTM architecture only train the datasets from left-to-right sequence, which uses the right context, but neglected the left-to-right sequences. Bi-directional LSTM (Bi-LSTM) architecture is able to train the network using both sequences and concatenate the two separate outputs to form produce one final results.

Therefore, we utilize a Bi-LSTM RNN-based model to classify responses into one of the two categories: positive and negative. In this section, we will introduce the principles of both LSTM and Bi-LSTM and proposed our model for mood detection.

LSTM differs from typical RNN networks; it uses an input $i_t$, output $o_t$, and forget gate $f_t$, as well as cell state $C_t$. In order to maximize our model performance, we adopt a popular LSTM variant proposed by Alex Graves (Graves, 2013), which prevent the gradient from being too large. The formulas are listed as follows:

$$i_t = \sigma(x_t W_{xi} + h_{t-1} W_{hi} + W_{ci} \odot c_{t-1} + b_i) \tag{2}$$

$$f_t = \sigma\left(x_t W_{xf} + h_{t-1} W_{hf} + W_{cf} \odot c_{t-1} + b_f\right) \tag{3}$$

$$o_t = \sigma(x_t W_{xo} + h_{t-1} W_{ho} + W_{co} \odot c_t + b_o) \tag{4}$$

$$C_t = f_t \odot C_{t-1} + i_t \odot \sigma(x_t W_{xc} + h_{t-1} W_{hc} + b_c) \tag{5}$$

$$h_t = o_t \odot \sigma(C_t) \tag{6}$$

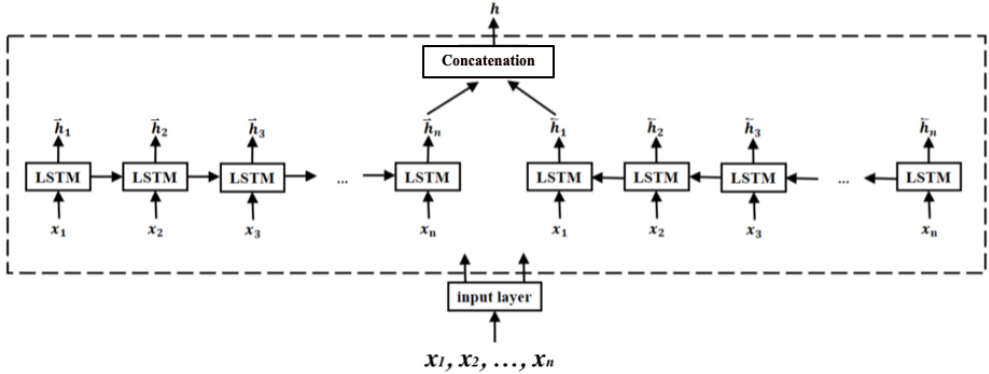

*Figure 4.2:* The Architecture of Bi-LSTM

where $\sigma$ denotes the sigmoid activation function. The weight matrixes could be understood as the subscripts suggested; for example, the $W_{ho}$ is the hidden-output gate weight matrix. Finally, the output of the LSTM network is the hidden state as shown in (6). Overall, the architecture of our model is shown in figure 4.2.

Bi-LSTM had proved to be a more effective model than LSTM as it is train by using two sequences (Chen, 2017), which brings out a more comprehensive picture of the context. In our approach, we are able to train the given input from two sequences: one from normal sequences, and another from backward sequences. We use $x$ to denote the input sentence and use $h$ to denote the feature vector of the sequences, which contains the context of the current word and the words before the current word. Because we have both directions, we will get a vector from the normal sequences $\vec{h}$ and another from the reverse sequences $\overset{\sim}{h}$. In order to get the final output for our Bi-LSTM network, we concatenate the two vectors into $h$. Finally, we attain the results by activating the hidden state through sigmoid function. The formulas are as shown below:

$$h = [\vec{h}, \overset{\sim}{h}] \tag{7}$$

$$y = \sigma(h) \tag{8}$$

As shown in figure 4.3, our binary classification model contains three parts: the input layer, Bi-LSTM network, and sigmoid activation function. When words are fed to the model, they are transformed into embedding vectors using a trained word embedding model. These will serve as the input layer that will be fed the Bi-LSTM network. After the training, the output vector will go through the activation function sigmoid and produce a probability from 0 to 1, where 0.5>output will be labeled as negative, and output>=0.5 will be labeled as positive.

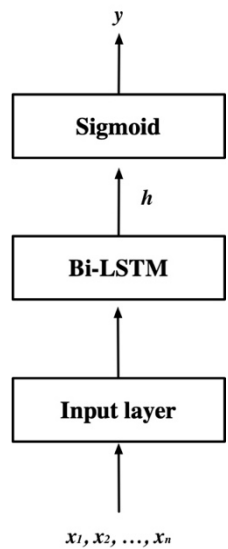

*Figure 4.3:* The Sentiment Analysis Model

| Algorithm 1: **Sentiment analysis** |
|---|

1.   Pos, neg, train: elements in the corresponding list
2.   
3.   Load pretrained word embedding model cn_model
4.   **Input**: list of positive text and negative text, pos_txts and neg_txts respectively.
5.   **For** pos **in** pos_txts.length
6.       append pos in train_txt
7.   **End For**
8.   **For** neg **in** neg_txts.length
9.       append neg in train_txt
10.  **End For**
11.  //tokenize the text
12.  **For** train **in** train_txt.length
13.      Remove any punctuation from train
14.      cut train into words <- cuts
15.      Append the cuts into cuts_list
16.      **For** cuts **in** cuts_list
17.              tokenize cuts <- tokens
18.              Append tokens into tokens_list
19.      **End For**
20.  **End For**
21.  //training session
22.  train_pad = padding and truncating tokens_list into equalized length
23.  **train** Bi-LSTM using train_pad for 20 epochs using cross entropy loss
24.  **Output**: a probability value using sigmoid activation function

## 4.4 Objective Function

Since we need the model only to classify between negative and positive, we set the model's goal as a binary classification task, which will use the binary cross entropy loss function to calculate the loss:

$$BCE = \frac{1}{N}\sum_{i=1}^{N} -(y_i log(p(y_i)) + (1 - y_i)log(1 - p(y_i))) \tag{9}$$

Where y is the ground truth label (1 for positive and 0 for negative) and p(y) is the predicted label given a dataset of N numbers of input.

## 4.5 Experiments & Results

### 4.5.1 Dataset and Data Preprocessing

As neural networks cannot accept sentences as input, we must first convert the data into a format available for the network to process. Therefore, we utilize pretrained word vectors available on GitHub to separate all data into individual words and convert each word into an index corresponding to an entry in the pretrained word vector.

We store all the data in a long Python list, which consists of positive emotion responses followed by negative emotion responses. Then we convert all data into word tokens. In order to achieve this, we have to remove all punctuation present in the data, then use the Python module Jieba to separate data into words. As shown in equation 10, each entry can be in a form of an array of index numbers, corresponding to words in the pretrained word vector.

$$X_{tk} \in \{E_1, E_2, \ldots, E_i\} \tag{10}$$

$$E_i = (S_i, Etype_{Si})$$

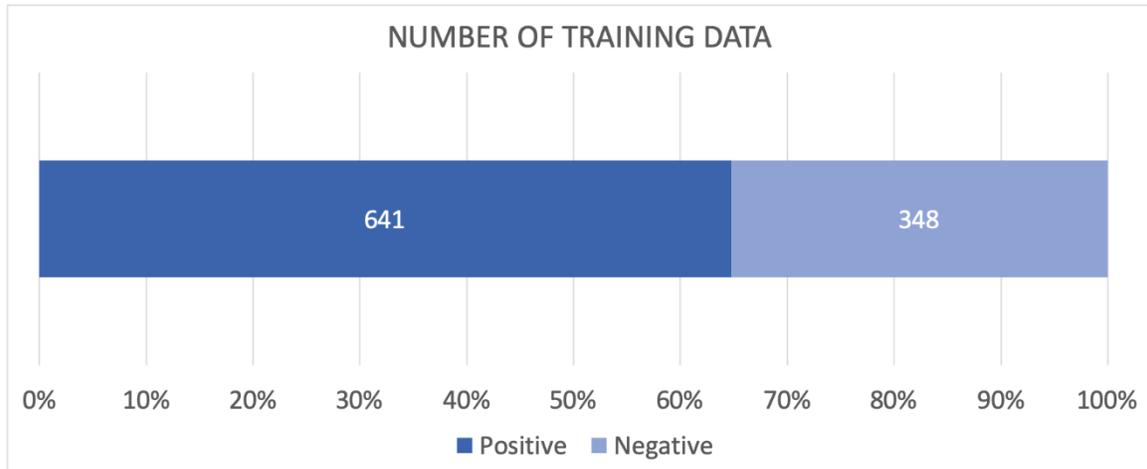

*Figure 4.4:* Proportion of Training Data with Different Labels

Figure 4.4 shows the proportion of negative and positive training data extracted from the questionnaire. The results are in accordance with our assumption that responses to each question would more often be positive responses rather than negative responses, because at a given time in the population, only a small portion of people would possess negative emotion. Therefore, we have sufficient evidence that we have collected a set of data that is suitable for training.

The model can only accept vectors of a restricted length, but the length of each entry is different. If we simply select the longest entry and pad the others into the same length, we would waste computational resources. Therefore, we select a compromising length that will cover 95 percent of the data, cutting the longer entries and padding the shorter ones in an iteration. In this way, all entries can be of the same length. We use the 'pre' padding because past experiments have shown that if we fill zeros after the indices, potential unwanted effects could appear in the model.

Now we can prepare an embedding matrix for the model. We need to prepare a matrix of dimension (NumWords, EmbeddingDim), with NumWords representing the number of words we use and EmbeddingDim representing the length of the vector representing each word, as required by documents of Keras[2], our framework for building the model.

As one of our goals is to conserve calculational resources, we only utilize 100,000 most used words in a total of about 260 thousand words inside the pretrained word vector. This is possible as we have only a small sample of size 990, and most of the words in the sample are common words rather than rare or archaic ones.

---

[2] Keras documentation: https://keras.io/layers/embeddings/

### 4.5.2 Model Implementation

We implement the model using Keras with TensorFlow backend, in a single Jupyter Notebook with Python3 kernel. This setup makes the model portable and executable on a wide range of systems, from lightweight mobile workstations to GPU-assisted servers. Keras also contains all library code we need to construct our model, saving the need to look for optimal functions in different code libraries.

When we train the models, we use the optimizer Adam with a small learning rate to prevent skipping over the optimal solution in gradient descent as a result of large learning rates. We monitor the value of validation loss while training, and use the functions Early Stopping and Reduce Learning Rate on Plateau based on the value to alleviate overfitting in the model. We also save model checkpoints at the end of every epoch to avoid data loss in sudden unpredictable scenarios.

### 4.5.3 Hyperparameter Settings

In our model, the learning rate is initially set to 0.001 as smaller learning rates benefit the model in that the plateau may be found earlier. The Early Stopping callback function monitoring the value of validation loss is set to stop model training when the loss does not improve in 3 epochs, and the Reduce Learning Rate on Plateau function slows the learning rate by a factor of 0.1 every epoch when the validation loss does not improve, eventually reaching a learning rate of 0.00001. In terms of the architecture of the network, we have two intermediate layers, a Bi-LSTM layer and an LSTM layer, with 32*2 and 16 hidden units respectively for each layer.

### 4.5.4 Results

We deployed a single workstation with a dual-core CPU and 4 gigabytes of RAM to train the model on our collected training data, as the model is not intense on computational resources and could run efficiently even on low-end machines. After training, we test the model on the test set. The experimental results are shown in Table 1 and compared to multiple other solutions.

We use the $F_1$-score metric, which is the harmonic average of the precision and recall, on each model. The $F_1$-score is calculated as:

$$F_1 = \left( \frac{recall^{-1} + precision^{-1}}{2} \right)^{-1} = 2 \cdot \frac{precision \cdot recall}{precision + recall} \tag{11}$$

| Model | Precision of test set | $F_1$-score |
|---|---|---|
| **GRU (Baseline)** | 87.74% | 0.846 |
| **LSTM** | 89.26% | 0.884 |
| **Bi-LSTM** | 90.91% | 0.928 |

*Table 4.2:* Experimental comparison of models

It can be observed from Table 4.2 that:

- The Bi-LSTM model has a considerable advantage over the baseline, achieving 3.17% more in precision of test set and a higher $F_1$-score than GRU. This matches our expectation that more complex representations results in more accurate predictions, given that overfitting does not occur.
- The LSTM model performs similarly with the Bi-LSTM model but with slightly lower precision and $F_1$-score. This shows that a Bidirectional layer can help when natural language processing is involved.

Examples of classifications on the test data is shown below in Table 4.3 for comparison:

| **True Positives (Predicted = Positive, Actual = Positive)** |
|---|
| 我一般都在和同学聊家常去吃什么去玩什么各种 |
| (My classmates and I usually talk about what to eat, what to play and other trivia) |
| 我希望他能改变自己的看法但我会先听他的意见 |
| (I hope he can change his views, but I will first listen to his comments) |
| 谢谢并接受 |
| (Thank [him/her] and accept it) |

| **True Negatives (Predicted = Negative, Actual = Negative)** |
|---|
| 当做没看到 |
| (Pretend not to see it) |
| 买那个挑衅的因为另外一个有可能是假票 |
| (Buy [the ticket] from the provocative person as the other person may be selling fakes) |
| 虚假的 |
| ([I think it] is fake) |

| **False Positives (Predicted = Positive, Actual = Negative)** |
|---|
| 我会再看看别的公司 |
| (I will check out other companies) |
| 我会买人的票 |
| (I will buy someone's ticket) |
| 弄备份瞎玩儿 |
| (Create backup and play with it) |

| **False Negatives (Predicted = Negative, Actual = Positive)** |
|---|
| 我爱我 逃 |
| (I love my escape) |
| 世界上某个人的人生毕竟梦可以预言 |
| ([It should be] someone in the world's life as dreams can prognosticate) |
| 学习 |
| ([I will] study) |

*Table 4.3:* Examples of classifications on the test data

It can be seen that if the meaning of a phrase or sentence is clear, the model can accurately predict the polarity of that phrase or sentence. However, if the meaning is vague or obscure, or that the emotion in the sentence is considered neutral, the model may not classify correctly. Note that the model classifies doubting or cynical emotions to be negative as we may consider these as evidence of stress, depression or anxiety.

## 5 Retrieve Counseling-related Corpora

### 5.1 Problem Definition

Deep learning based chatbots' performance not only depends on its training architecture, but also on the Q&A pair corpus, or its dataset, that is used for the training. Generally, increasing the size of the training dataset will lead to an increase in performance of the model. A massive dataset is therefore required to train an acceptable conversational robot. However, because of the privacy issues specific to the field of psychological counselling and illness, there are no available open datasets on the internet. There is also a lack of sources to obtain a direct dataset for psychological related corpora that is suitable for training. In this chapter, we propose our approach to address this issue.

For large datasets, it is very hard to annotate data manually, as it would require tremendous amounts of labor resources and time. Our approach to this issue is suggested as followed:

1. Manually annotate a small number of psychological-related datasets from a qualified source, then use it to train a classifier that can classify between whether it is psychological related or not related.
2. Apply the trained classifier on large datasets, filter out those that are unrelated and only obtain those that are psychological related.

After the filtering process, we could use the filtered datasets to train our response model.

### 5.2 Data Analysis

In order to select sources qualified for training, we find multiple college forums, specifically from Tsinghua and Peking University, and look for topics and posts related to mental healthiness, where students convey their true emotions to the online therapist. This data is valuable as most of the information is psychological-oriented, which is close to our research topic on campus therapy.

The analysis of the data is shown in the figure 5.1. We have collected nearly 30% of the data that are mentally related, and 70% of the data labeled as casual (mentally unrelated). This data will be then fed into the model that can determine the type of the data itself, minimizing the time cost for labor.

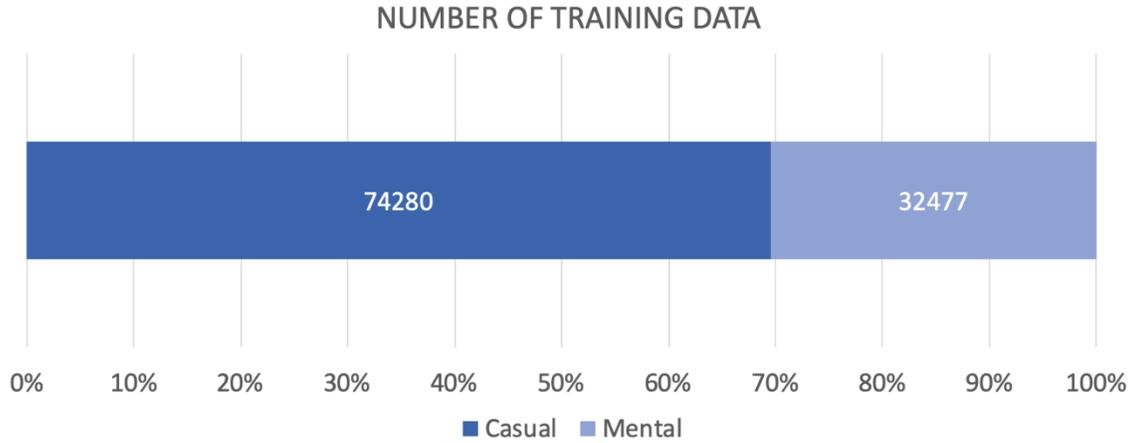

*Figure 5.1:* Proportion between Casual and Mental Training Data

### 5.3 Deep Learning-based Text classifier

The model's flowchart is shown on the figure below: Using a trained classifier $F$ and an evaluate function $E(x)$ through a set of annotated data $D_a$, we could attain a set of psychological-related data $D_r$ in the set of un-annotated data $D_u$.

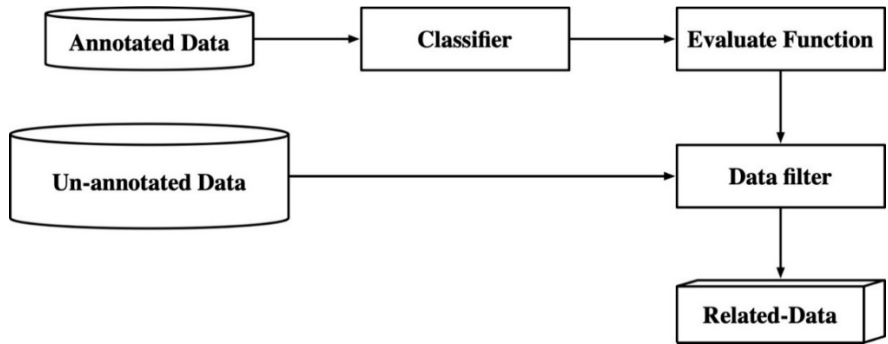

*Figure 5.2:* Mental-related Corpora Classifying Model

Since the model's task is a binary classification task that is similar to section 4, we will use the already proposed Bi-LSTM model for classifying the corpora. The differences are in actual implementation of the model, which will be discussed in section 5.4. Models objective are shown in formula(number). Specifically, we choose binary cross entropy loss, denoted as $BCE^{Class}$ in Equation 12, as our loss function.

$$BCE^{Class} = \frac{1}{N}\sum_{i=1}^{N} -(s_i log(p(s_i)) + (1 - s_i)log(1 - p(s_i))) \tag{12}$$

Algorithm 2: **Classifier – classifying process**

```
1.    isMentalRelated :  judge if the dialogue relevant to Mental consult
2.    followingLines : a list to store Mental dialogue
3.    input: open the document 'corpora.conv' which store dialogue information
4.    line <- the contents in each line of 'corpora.conv'
5.    while line is not null
6.        begin
7.        if line[0]='E'
8.                position <- the path of the document
9.            isMentalRelated <- false
10.        entry <- the content of next line
11.        while entry is not 'E'
12.            begin
13.                position <- the path of the document
14.                append entry into followingLines
15.            if the relevant of entry >= 0.7
16.                begin
17.                        isMental <- true
18.                        entry <- the content of next line
19.                for item in followingLines
20.                    write 'E' in to 'corpora_Mental_related_corpora.txt'
21.                      print a line break
22.                    for item in followingLines
23.                    output variable item and a line break into'Mental_related_corpora.txt'
24.                    renovate in 'corpora_Mental_related_corpora.txt'
25.                        End For
26.                End For
27.                End
28.            End
29.    Output: corpora_Mental_related_result.txt
```

## 5.4 Results & Experiment

### 5.4.1 Preprocessing

We have already introduced in section 4.5.1 the details of preprocessing the data. The training datasets only included those data which can successfully converted to vector form for training. As shown in figure 5.1 above, we prepared 74280 casual data and 32477 mental related data after preprocessing, among which we will randomly set 90% as training data and 10% as test data.

Then we start the data cleansing, removing any punctuation, alphabet, and numbers, filter out sentences with extreme length, and delete any non-Chinese character. This will serve as our datasets for final training.

### 5.4.2 Model Implementation

Similar to section 4.5.2, we implement the model using Keras with TensorFlow backend, in a single Jupyter Notebook with Python3 kernel. This setup makes the model portable and executable on a wide range of systems, from lightweight mobile workstations to GPU-assisted servers. We utilize the same workstation to train the model.

When we train the models, we use the optimizer Adam with a learning rate of 0.001 to 0.00001, monitor the value of validation loss while training, and use the functions Early Stopping and Reduce Learning Rate on Plateau based on the value to alleviate overfitting in the model. We also save model checkpoints at the end of every epoch to avoid data loss in sudden unpredictable scenarios.

In terms of the architecture of the network, we still utilize two intermediate layers, a Bi-LSTM layer and an LSTM layer, with 32*2 and 16 hidden units respectively for each layer.

### 5.4.3 Analysis of the results

The testing of the model on the test set shows experimental results summarized in Table 2 and compared to multiple other models. We use the precision of test set and $F_1$-score metrics in the comparison.

| Model | Precision of test set | $F_1$-score |
|-------|----------------------|-------------|
| GRU (Baseline) | 84.29% | 0.801 |
| LSTM | 87.56% | 0.824 |
| Bi-LSTM | 90.37% | 0.833 |

*Table 5.1:* Examples of classifications on the test data

It can be observed from Table 5.1 that:

- The Bi-LSTM model still retained its advantage over the baseline, achieving 6.08% more in precision of test set and a 0.32 higher $F_1$-score than GRU. This matches our expectation that more complex representations results in more accurate predictions.
- The LSTM model performs similarly with the Bi-LSTM model but with slightly lower precision and $F_1$-score, consistent with our previous review. This shows that a Bidirectional layer can help when natural language processing is involved.
- With a much larger dataset compared to our mood detection model, the classifier instead scored a lower overall $F_1$-score. This may be attributed to a larger change of misclassification in a larger dataset or may be considered a result of the subjective nature of psychological counseling relatedness.

Examples of classifications on the test data is shown below in table 5.2 for comparison.

| True Positives (Predicted = Related, Actual = Related) |
|---|
| 所以舍弃任何一方都非常不容易 |
| (So giving up either side is extremely hard) |
| 当时还没有 这个恶性循环的想象中 |
| (At that time [I] did not, in imagination of this vicious cycle…) |
| 渐渐地就麻木了 |
| ([I] gradually became apathetic) |

| True Negatives (Predicted = Unrelated, Actual = Unrelated) |
| --- |
| 中新网 3 月 1 日电 Twins 门前 澳门合体为《英皇盛世 10 周年巨星演唱会》压轴演出 |
| (China News, 1ST March, in front of Twins gate: Macao unites to present final performance for the 10th Anniversary Royal British Superstar Concert) |
| 把放映机的灯泡人为替换 |
| ([To] artificially replace the light bulb in the projector) |
| 前两天 只身赴福冈旅游 4 天 3 夜散心 |
| (Lately [I] travelled to Fukuoka alone for 4 days and 3 nights to relieve boredom) |
| **False Positives (Predicted = Related, Actual = Unrelated)** |
| 再另挑选自己喜欢的课程 |
| (To pick another course that I like) |
| 如果是在现代碰到这种情况 |
| (If [I] see this happening in modern times) |
| 自然会摒弃深刻的理性 |
| (Nature will abandon profound rationality) |
| **False Negatives (Predicted = Unrelated, Actual = Related)** |
| 爷爷走得早 |
| (Grandfather passed away early) |
| 我想知道 |
| (I would like to know) |
| 目标不够明确 |
| (The goal is not clear enough) |

*Table 5.2:* Examples of classifications on the test data

It can be seen in Table 5.2 that if a phrase or sentence is clearly psychological counseling related or unrelated, the model can accurately predict that phrase or sentence to be related or unrelated. However, as it is inherently subjective in judging if a sentence is psychological counseling related, and sentences themselves are often vague or obscure, the model may not classify correctly at all times. Looking at the falsely classified examples, we note that many of these we can consider to be correctly classified as well due to their ambivalent nature.

**5.5 Filtering the Collected Corpora**

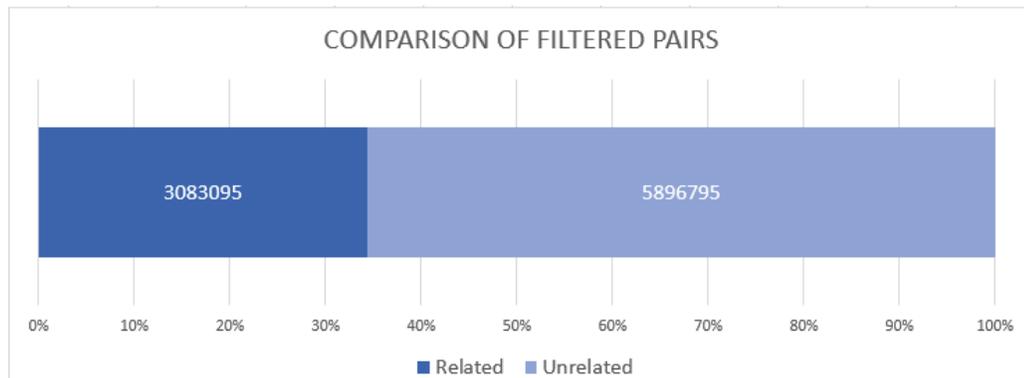

*Figure 5.2:* Proportion between Filtered Q&A Pairs in Corpora

After we had trained our classification model, we were able to classify our collected corpora into the two categories, either psychological counseling related or unrelated. When filtering through the corpora, we retain the entire Q&A pair when either the question or the answer is psychological counseling related. In this way we retain the Q&A pair structure of the original corpora, and avoid the problem of having an isolated question or answer.

In a total of nearly 9 million Q&A pairs collected, our model filtered out about 3.1 million related to psychological counseling while classifying the other 5.9 million unrelated, as shown in figure 5.2. This coincides with our expectation that most of the discourse on public forums would not be psychological counseling related, as people tend not to talk about their own mental problems in public areas such as forums on the internet.

The automatic classification of large corpora using the classifier has greatly reduced the labor and temporal costs of data filtering, and we could therefore proceed to building the response model in a relatively short amount of time.

## 6 Response Models

### 6.1 Introduction

Our response models are generation based, which can generate answers according to user's responses. An effective response model should be able to produce fluent, grammatical, and meaningful sentences. But in reality, most of the generation-based model tend to output trivial and redundant responses like "You are a good man" or "I don't know" (Serban et al, 2016). This can be explained in terms of the frequency of the data in datasets. Whereas responses like "I don't know" may appear very frequently in the dataset, more meaningful responses appear to be in a sparse situation. To solve this issue, we built our Seq2seq model based on an objective function Maximum Mutual Information (MMI) as an anti-language structure (Li, 2016).

### 6.2 Seq2Seq generation model

Given a sequence of inputs $X = \{x_1, x_2, \dots x_n\}$, we want to predict a sequence of outputs $Y = \{y_1, y_2, \dots y_n\}$. Equation 13 gives the conditional probability of generating the outputs Y that contains nth $y$, where $f$ denotes a nonlinear activation function, $e_{yk}$ denotes single units at time step k in vector form, and $h_{k-1}$ is the representation output at time k-1(Li, 2016).

$$p(Y|X) = \prod_{k=1}^{n} p(y_k \,|x_1, x_2, \dots x_n \,, y_1, y_2, \dots y_{k-1})$$
$$= \prod_{k=1}^{n} \frac{exp(f(h_{k-1}, e_{yk}))}{\sum_{y'} exp(f(h_{k-1}, e\ y\ '))} \tag{13}$$

RNN units: Because we are dealing with relatively long sequence datasets, we use LSTM as our RNN unit as it could contain more long-range relations during training.

Decode: During in the decoding phase, we utilize beam search with beam size of 5 as opposed to greedy search. Choosing beam search allow us to be less likely to accumulate error in the decoding process.

### 6.3 MMI model

First used in speech recognition (Bahl, 1986), MMI are a method that evaluate the mutual dependence between inputs and output. Li et al. successfully implement this method into response generation task. Let $S$ be an input message sequence $S = \{s_1, s_2, \dots s_n\}$, where word $s_n$ is in position $n$. $T$ denotes the corresponding output response sequence $T = \{t_1, t_2, \dots t_n\}$. Each sentence contains a special token EOS, which denotes the termination of the algorithm. Equation 14 show the objective function of MMI, where parameters are dedicated to maximizing the mutual relationship between S and T:

$$\log \frac{p(S,T)}{p(S)p(T)} \tag{14}$$

Using this equation, high frequency responses are less likely to be favored than do another model. Instead, MMI model will favored to responses that are sparse but aim specifically at a given input. The objective function with a hyperparameter $\lambda$ is suggested as follows:

$$T = \arg \max\{\log p(T|S) - \lambda \log U(T)\} \tag{15}$$

where p(T) and U(t) are language models that penalizes the given inputs and decide the output of the model.

## 6.4 Response Strategies

In order to fully utilize the potential of our chatbot model and to improve performance in response generalization altogether, we designed a response strategy that can switch between the casual talk chatbot and the psychological counseling chatbot to improve the performance of chatbot and experience of the user.

Upon receiving input from the user on the online platform, we first use the mental or casual classifier to decide if the user's discourse concerns psychological counseling. If the input is deemed casual, we use the casual talk chatbot for the user. If the input is deemed psychological, we further utilize the positive or negative classifier to determine the polarity of the discourse. If the input is deemed positive, we fall back to the casual chatbot; if the input is deemed negative, we can be determined stress or anxiety is present and use the psychological counseling chatbot.

Regardless of the chatbot used, we track the decimal score output of each query through both classifiers. If there is a trend towards negative emotion or need of psychological counseling, as indicated by the moving average of past 5 inputs, we shift the chatbot used to the psychological counseling one if it is not already so. If there is a trend towards positive emotion or no longer a need for psychological counseling, as indicated by the same moving average, we shift the chatbot used to the casual talk one instead.

## 6.5 Experiment

To evaluate the effectiveness of our chatbot, we developed a scoring system $R_{check}$ for testing. Because of the lack of chatbots having similar functions, our only method is to use subjective evaluation to carry forth the experiment.

For evaluation, we have prospective students analyze the responses produced by our response model (five analysts looking through the responses in chatbot), and score each response with three separate levels: "0" means the response is unqualified, "1" means the response is regular, and "2" means the response is qualified. To be more specific in how to determine each level, we establish three metrics: 1. Grammatically correct, 2. Topic related, 3. Not too vague. The chatbot's response will only be label regular if the previous two are fulfilled. If all three metrics are fulfilled, the response will be label as qualified.

In order to have a quantitative view on the results, we developed $R_{check}$ as an index, as shown in equation below.

$$R_{check} = (N_{reg} + N_{qualified})/ N_{total} \qquad (16)$$

### 6.5.1 Evaluating the model

We conducted a randomized experiment in school campus with 200 random sampled questions. We show the $R_{check}$ scoring system result in Table 6.1, which suggest the $R_{check}$ index and the proportion of each label.

| $N_{unqualified}$ | $N_{reg}$ | $N_{qualified}$ | $R_{check}$ |
|---|---|---|---|
| 0.09 | **0.52** | 0.39 | 0.91 |

*Table 6.1:* Quantitative Evaluation of Response Model

As shown in table 6.1, the proportion of the $N_{reg}$ constitute up to half of the proportion of all the response, and $N_{qualified}$ reach to nearly 40%. This result is proven to be acceptable as the response are generally satisfying, as only 9% of the times the response is unqualified.

### 6.5.2 Online Counseling

After we have evaluated our model, we launched Evebot inside a server, and developed an online service on the platform WeChat. The platform allows our response model to chat with any user (primarily students) who intended to. The convenience and anonymity that the platform brings help our system to enhance its effectiveness. As user don't need to worry about the anxiety and privacy during in face to face contact, they are more willing to open their emotions toward our chatbot. We show an example of conversation generated by our response model in figure 6.1 and translated version in table 6.2.

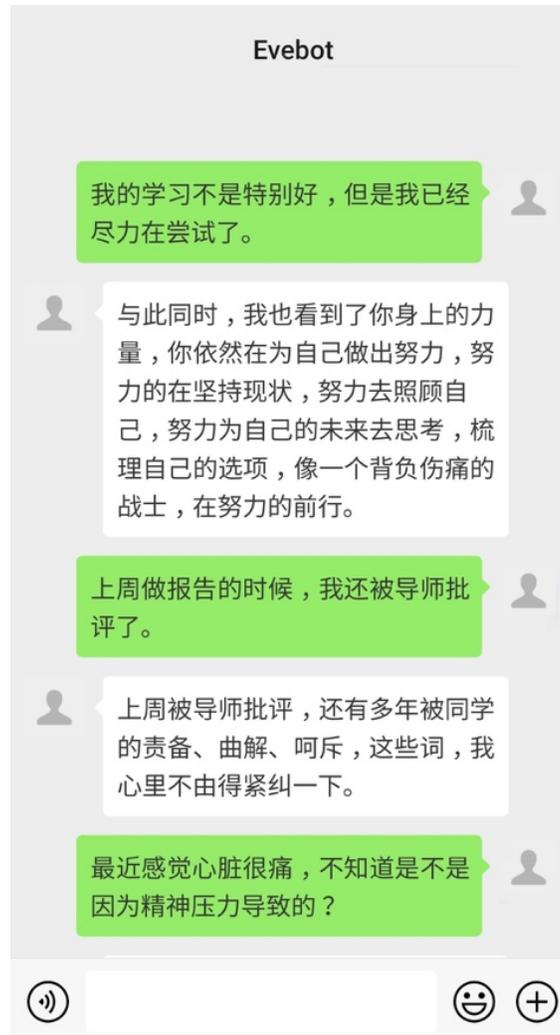

*Figure 6.1:* User interface of the response model

| **Conversation:** translated dialogue between response model and user |
| --- |
| **User:** I am not good at studying, but I did try my best. |
| **Evebot:** At the same time, I see diligence reside in your body. You are still making an effort for yourself, making an effort for maintain the situation, making an effort to self-care, making an effort to think about the future, trying to filter your choices, like an injured warrior, striving to move forward. |
| **User:** Last week while I made a presentation, I got criticized by my mentor. |
| **Evebot:** You get criticized by mentor, and you may also receive rebuke by your classmates, [hearing] these words, I feel very sorry for you in my heart. |

*Table 6.2:* Translation of the user interface responses

# 7 User Study

## 7.1 Experimental Design

We want to investigate whether the chatbot significantly improves the moods of its users in a statistical way. Therefore, we define our null and alternative hypotheses below, where $\mu_D$ represents the true mean difference between the mean improvement in positivity score (as ranked by our sentiment analysis classifier) using our Evebot $\mu_{Evebot}$ and the mean improvement in positivity score using a regular chatbot which does not have particularly mood alleviating properties $\mu_{Regular}$, i.e. $\mu_D = \mu_{Evebot} - \mu_{Regular}$.

$$H_0: \mu_D = 0$$

$$H_a: \mu_D > 0$$

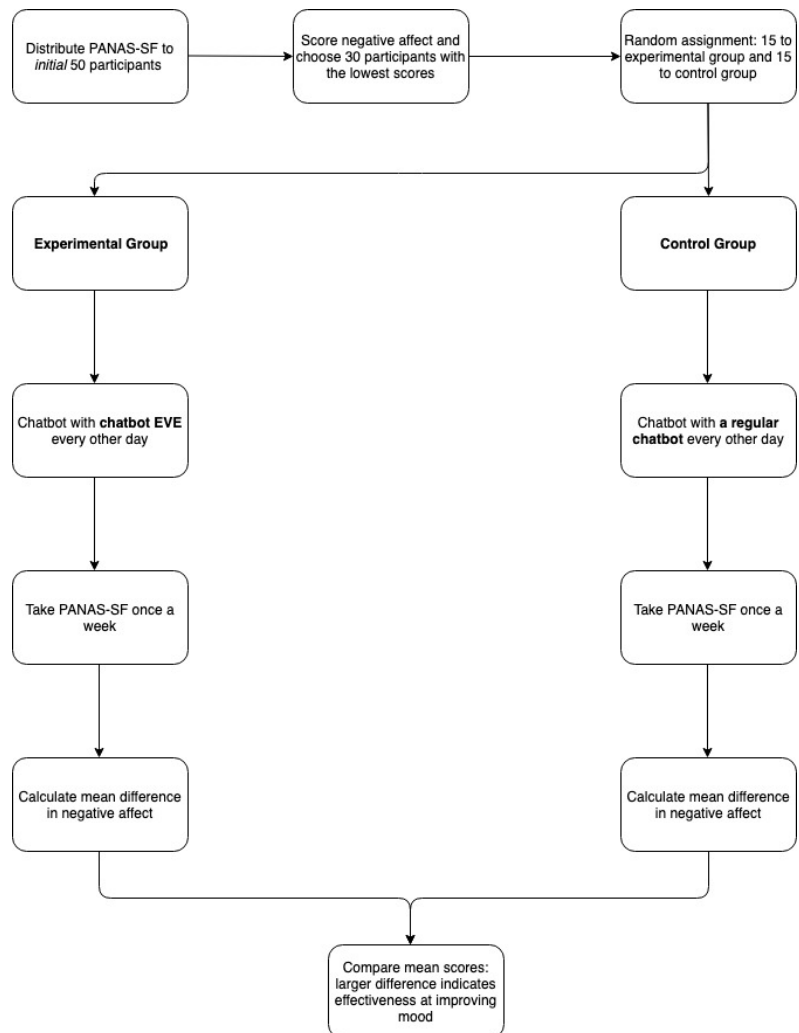

*Table 7.1:* Overview of User Experimental Study Process

We randomly select 50 initial participants in our high school. These participants are students and are fairly good representatives of the current adolescent population as they are all in their teenage years. We first gave out 50 copies of the PANAS-SF (Watson, D., Clark, L. A., & Tellegen, A.,1988) survey to those participants as our pre-test. Then, we collected the filled surveys and scored them according to the scoring scale given. Since we are looking at how Evebot improves an individual's mood, we use negative affect scores only.

After analyzing the results, we selected 30 surveys with the most negative affect and asked those 30 individuals to participate in the next stage of the experiment. Upon receiving informed consent, we randomly assign the participants to the experimental group and control group, with 15 participants in each group. The independent variable of this experiment is the type of chatbot the participants interact with, and other confounding variables such as time differences are kept controlled by asking the participants to complete their task on proposed days.

The experimental group is presented with Evebot, while the control group is presented with a regular chatbot, which does not have particularly mood alleviating properties. Participants of each group were asked to chat with their respective chatbots every other day, over the course of one month.

Following this, we gave out the PANAS-SF test once again to all of the participants, as our post-test. We collected the tests and obtained the scores for negative affect. Then, we compared each individual's scores between the pre-test and post-test and calculated the difference in the scores. We then calculated the mean improvement of the scores of the experimental group and the control group. In this case, the decrease in the score of negative affect indicates an improvement in mood and general psychological health. If the mean improvement of the experimental group is significantly higher than of the control group, then it can be said that the use of Evebot will improve the moods of students.

## 7.2 Analysis of the results

We conducted the experiment with the 30 selected students over a course of one month and arrived at significant results. The 15 students in the treatment group chatting with Evebot all showed a significant improvement in mood, leading to a mean improvement in the PANAS-SF score of 14.3. The 15 other students in the control group chatting with the regular chatbot showed an overall improvement in mood of 8.6 on the PANAS-SF scale, which can be attributed to the effects of talking to any entity over time. The difference between the mean improvements of the two groups would be 5.7, amounting to a delta of more than the full score of a question on the survey. A graph comparing the two mean differences is shown below.

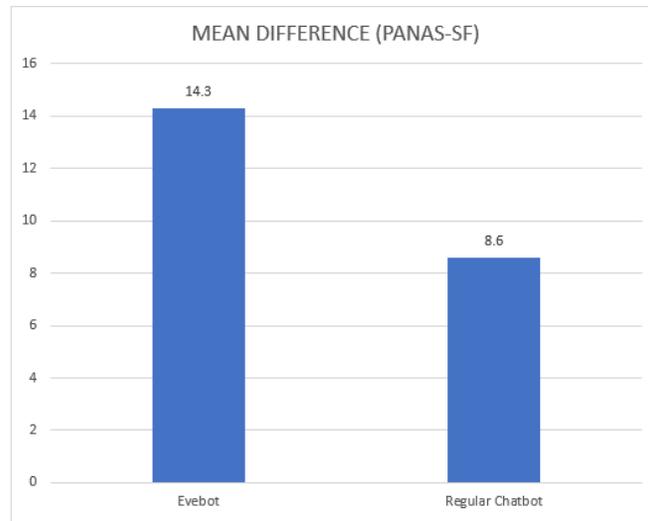

*Figure 7.1:* Comparison of Mean Differences between Evebot and Control

We have also tracked the performance of the 30 students using our Sentiment Analysis model. We take the arithmetic mean of all scores (as scored using the model) from two days of each user's input into the online chatbots and ordered the data into a line chart showing the score against time. It can be seen that an upward trend is evident in both the Evebot group and the Regular Chatbot control group, but a larger slope is clear in the Evebot group. Although there is only a negligible difference between the two groups in the first few days, the true difference quickly establishes over time.

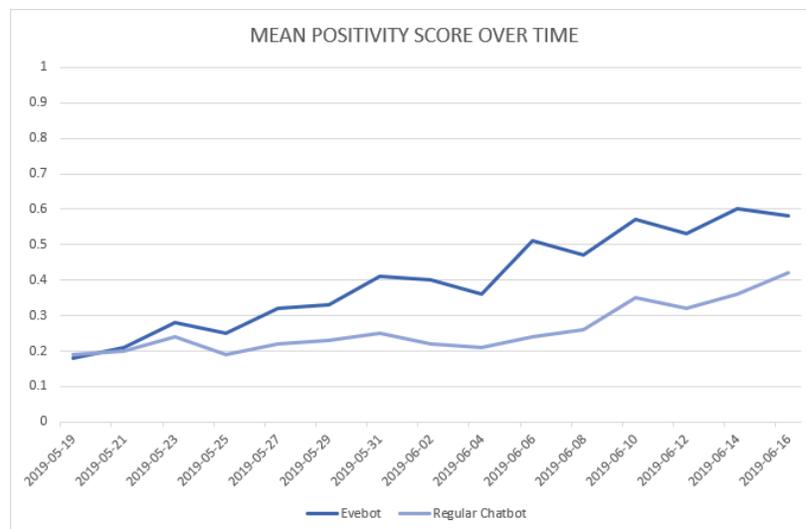

*Figure 7.2:* Positivity as Scored by Sentiment Analysis Classifier

Combining these two difference evidences, we can reject the null hypothesis. We have statistically significant evidence that there is a true difference in mood of the user between using the two chatbots, and that using Evebot is an effective way of relieving stress and becoming positive.

## 8 Conclusion & Expectations

### 8.1 Conclusion

In this paper, we propose an innovative conversational system and framework for diagnosing negative emotions and preventing depression through positive and suggestive responses. The resulting system will allow students to express their true feelings and therefore channel the negative emotion out, as both privacy and convenience are provided by the system. Our three proposed model consists of different machine learning based approaches, including Bi-LSTM model and Seq2seq model. Our contribution and innovative points are summarized as follow:

1. Proposed a new conversational system and framework capable of diagnosing emotions and chatting with the user at the same time, enhancing the experience of the user.

2. The proposed conversational system is fully generation based, which can generate new answers depend on each user question

3. Design a response strategy that can switch between the casual chatbot and the psychological counseling chatbot to improve the performance of the system and experience of the user

4. With a relatively small dataset of training, we still obtain a good result through the proposed classifier in section 5, proving our chatbot model is efficient

5. Both the evaluation of models and experimental user study show a good feedback of Evebot.

These innovations have enabled the user to gradually release and transform their negative emotions to positive ones. Because all of our models are deep learning based, as more interaction with the system occurred, more training data can be collected. Therefore, the performance of the system is likely to improve over time.

### 8.2 Expectations

There are also certain limitations in our system that we hoped to improve in the future. For our proposed conversation system, the limitations are summarized as follow:

1. As suggested in the evaluation of our response model, there is still about 10% of the responses are not qualified for an effective conversation. The grammar and meaning of the responses may performed poorly if user's question is out of the scope of our training data.

2. The conversational system is incapable of "remembering" the previous conversations. This means that during in the conversations, if the user is talking about the same topic for two separate questions, our conversational system will not remember what it generated previously in the conversation. Therefore, the answer generated corresponding to each question may be very different, which can confuse the user.

In our future work, we could fetch and collect more data using the classifier and the system to enlarge the scope of our system response. We could also add some elements of rule based chatbot to allow the chatbot to "remember" its responses.

# 9 References

Bahl, Lalit R., et al. "Maximum mutual information estimation of hidden Markov model parameters for speech recognition." *proc. icassp.* Vol. 86. 1986.

Balahur, Alexandra, Jesús M. Hermida, and Andrés Montoyo. "Detecting implicit expressions of sentiment in text based on commonsense knowledge." *Proceedings of the 2nd Workshop on Computational Approaches to Subjectivity and Sentiment Analysis.* Association for Computational Linguistics, 2011.

Chen, Tao, et al. "Improving sentiment analysis via sentence type classification using BiLSTM-CRF and CNN." *Expert Systems with Applications* 72 (2017): 221-230.

Compas, Bruce E., Pamela G. Orosan, and Kathryn E. Grant. "Adolescent stress and coping: Implications for psychopathology during adolescence." *Journal of adolescence* 16.3 (1993): 331-349.

Graves, Alex. "Generating sequences with recurrent neural networks." *arXiv preprint arXiv:1308.0850* (2013).

Hawton, Keith, Kate EA Saunders, and Rory C. O'Connor. "Self-harm and suicide in adolescents." *The Lancet* 379.9834 (2012): 2373-2382.

Li, Jiwei, et al. "A diversity-promoting objective function for neural conversation models." *arXiv preprint arXiv:1510.03055*(2015).

Liao SH. Expert system methodologies and applications – A decade review from 1995 to 2004. Expert systems with applications. 2005; 28(1), 93-103.

Lin, Jimmy, and Boris Katz. "Question answering from the web using knowledge annotation and knowledge mining techniques*." Proceedings of the twelfth international conference on Information and knowledge management.* ACM, 2003.

Liu, Yuanchao, et al. "Pal: a chatterbot system for answering domain-specific questions." Proceedings of the 51st Annual Meeting of the Association for Computational Linguistics: System Demonstrations. 2013.

Rogers, Carl Ransom. On becoming a person: A therapist's view of psychotherapy. Houghton Mifflin Harcourt, 1995.

Serban, Iulian V., et al. "Building end-to-end dialogue systems using generative hierarchical neural network models." *Thirtieth AAAI Conference on Artificial Intelligence.* 2016.Serban, Iulian V., et al. "Building end-to-end dialogue systems using generative hierarchical neural network models." *Thirtieth AAAI Conference on Artificial Intelligence. 2016.*

Shaw, Lindsay H., and Larry M. Gant. "In defense of the Internet: The relationship between Internet communication and depression, loneliness, self-esteem, and perceived social support." *Internet Research* 28.3 (2004).

Subasic, Pero, and Alison Huettner. "Affect analysis of text using fuzzy semantic typing." *IEEE Transactions on Fuzzy systems* 9.4 (2001): 483-496.

Watson, D., Clark, L. A., & Tellegen, A. (1988). Development and validation of brief measures of positive and negative affect: the PANAS scales. Journal of personality and social psychology, 54(6), 1063.

Weizenbaum, Joseph. "ELIZA---a computer program for the study of natural language communication between man and machine." *Communications of the ACM* 9.1 (1966): 36-45.

Wills, Thomas Ashby, Grace McNamara, and Donato Vaccaro. "Parental education related to adolescent stress-coping and substance use: development of a mediational model." *Health Psychology* 14.5 (1995): 464.